\documentclass{article}
\usepackage[utf8]{inputenc}
\usepackage[T1]{fontenc}
\usepackage{hyperref}
\usepackage{spconf,amsmath,graphicx} % from icip example
\usepackage{psfrag,amsthm,amssymb}
\usepackage{url,color, xcolor}
\usepackage[hyperpageref]{backref}
\usepackage[caption=false]{subfig}
\usepackage{flushend}

\hypersetup{
colorlinks   = true, %Colours links instead of ugly boxes
urlcolor     = magenta, %Colour for external hyperlinks
linkcolor    = red, %Colour of internal links
citecolor   = {green!50!black} %Colour of citations
}

\usepackage{xspace}
\makeatletter
\DeclareRobustCommand\onedot{\futurelet\@let@token\@onedot}
\def\@onedot{\ifx\@let@token.\else.\null\fi\xspace}

\def\eg{\emph{e.g}\onedot} 
\def\ie{\emph{i.e}\onedot} 
 
\def\etc{\emph{etc}\onedot}

\makeatother

\usepackage{enumitem}
\setlist[itemize]{noitemsep, topsep=0pt}
\setlist[enumerate]{noitemsep, topsep=0pt}

\title{A metric for evaluating  3D reconstruction and mapping performance with no ground truthing}

\name{Guoxiang Zhang and YangQuan Chen}
\address{Mechatronics, Embedded Systems and Automation Lab,\\ University of California, Merced, Merced, USA}

\hypersetup{
 pdfauthor={Guoxiang Zhang and YangQuan Chen},
 pdftitle={A metric for evaluating  3D reconstruction and mapping performance with no ground truthing}}

\begin{document}

\maketitle
\begin{abstract}
It is not easy when evaluating 3D mapping performance because existing metrics require ground truth data that can only be collected with special instruments. In this paper, we propose a metric, dense map posterior (DMP), for this evaluation. It can work without any ground truth data. Instead, it calculates a comparable value, reflecting a map posterior probability, from dense point cloud observations. In our experiments, the proposed DMP is benchmarked against ground truth-based metrics. Results show that DMP can provide a similar evaluation capability. The proposed metric makes evaluating different methods more flexible and opens many new possibilities, such as self-supervised methods and more available datasets.
\end{abstract}


\section{Introduction}
\label{sec:org90f2610}

\textbf{Background and motivation.} Simultaneous localization and mapping (SLAM)~\cite{Cadena2016,MurArtal2017} and 3D reconstruction~\cite{Zhang2018aw,Choi2015,Garcea2018} methods have achieved dramatic advancements recently due to the increasing need for autonomous vehicles and consumer robots. But, it is still not an easy task to collect data for performance evaluation. 
Existing metrics require additional ground truth data, such as ground truth camera trajectories or ground truth 3D models, that need to be collected by special instruments, such as motion capture systems, RTK-GPS, and LiDAR.
These instruments are not only expensive. They make it impossible to add ground truth to existing data. As a result, there are only a limited number of datasets with ground truth information available~\cite{Firman2016}. 

To address this problem, some researchers turn to generate synthetic data from graphical rendering on virtual scenes~\cite{McCormac2017a,Song2017}. This makes accessing ground truth data easy, but it introduces new challenges to create realistic virtual environments and camera trajectories. Some datasets~\cite{Bayer2016,Dai2017b,Hua2016} choose to use pseudo-ground truth data which are generated by estimation algorithms, while some others resort to external markers~\cite{Kikkeri2014,Rogers2020}.

\noindent \textbf{Key contributions.} In this work, we propose a metric, dense map posterior (DMP), for 3D reconstruction and mapping performance evaluation that can work without any ground truth data. Instead, it calculates a comparable value, reflecting a map posterior probability, from dense point cloud observations. In our experiments, the proposed metric is benchmarked against ground truth-based metrics. Results show that the proposed DMP can provide a similar evaluation capability. 

\noindent \textbf{Significance.} The proposed metric makes a broad impact to SLAM and 3D mapping. 
We list a few major ones here:
\begin{enumerate}
\item It makes 3D mapping evaluation more flexible and easily accessible.
\item It provides a supervisory figure-of-merit signal for robust loop closure optimization.
\item It helps build large real-world SLAM datasets with minimal effort.
\item It makes it possible to introduce self-supervised machine learning algorithms to 3D reconstruction methods.
\end{enumerate}

\section{Related Work}
\label{sec:org4d9b570}
When evaluating 3D mapping results, the common error metrics are the absolute trajectory error (ATE)~\cite{MurArtal2017,Sturm2012,Whelan2015}, the relative position error (RPE)~\cite{Sturm2012}, and surface distance error~\cite{Zhang2018aw,Choi2015,Handa2014,Whelan2015}.
The ATE represents the difference between the ground truth and an estimated trajectory in a common world frame. The RPE calculates the relative relation differences between estimated and ground truth trajectories. 
The surface distance error specifies the mean and median of distances between a reconstructed surface and a ground-truth surface. 
All these metrics require ground truth data that is not trivial to collect.

The work closely related to ours is~\cite{Olson2009}. 
Olson and Kaess discussed using sparse map posteriors for map evaluation, but they concluded that the posterior subjects to over-fitting during the optimization process. Thus it is not reliable. In their formulation, the sparse map posterior is calculated on a map of sparse landmarks. We agree that, with sparse data only, it is not as reliable as ground truth-based metrics. However, in our method, we consider the posterior of a dense 3D map given geometry-related observations. The over-fitting is no longer a problem.

\section{Dense Map Posterior  (DMP)}
\label{sec:org6191777}
3D reconstruction and mapping methods usually take a set of sensor reading and output both an estimated 3D model (\eg mesh or surfels, \etc) $\mathcal{M}$ and an estimated camera or LiDAR trajectory $\mathbb{T}$. For simplicity, we denote $M=\{\mathcal{M}, \mathbb{T}\}$ to include both outputs. For the input sensor readings, we denote geometry related readings as \(\mathbb{Z}\), where \(\mathbb{Z}\) can be a set of depth images from RGB-D cameras or a set of point clouds from LiDARs. 

The idea behinds the proposed metric, dense map posterior (DMP), is to calculate a value reflecting a posterior probability \(p(M| \mathbb{Z})\). For this calculation, DMP takes both $M$ and $\mathbb{Z}$ as evaluation input. Then, performance can be evaluated by comparing metric values of different 3D reconstruction estimates \(\{M_i\}\).

Because there is no direct way to calculate \(p(M| \mathbb{Z})\), derivations are made to get a computable form. First, the Bayes' rule is applied to get 
\begin{equation}
\label{eq_post1}
p(M| \mathbb{Z}) = \frac{p(\mathbb{Z}|M)p(M)}{p(\mathbb{Z})}.
\end{equation}
Then, for the same set of observations \(\mathbb{Z}\), if there are two different 3D reconstruction estimates \(M_1\) and \(M_2\), we can get a ratio
\begin{equation}
\label{eq_lr1}
\frac{p(M_1| \mathbb{Z})}{p(M_2| \mathbb{Z})} = \frac{\frac{p(\mathbb{Z}|M_1)p(M_1)}{p(\mathbb{Z})}}{\frac{p(\mathbb{Z}|M_2)p(M_2)}{p(\mathbb{Z})}}.
\end{equation}
If we assume \(p(M_1)=p(M_2)\), we can simplify \eqref{eq_lr1} to be
\begin{equation}
\label{eq_lr2}
\frac{p(M_1| \mathbb{Z})}{p(M_2| \mathbb{Z})} = \frac{p(\mathbb{Z}|M_1)}{p(\mathbb{Z}|M_2)},
\end{equation}
where 
\begin{equation}
\label{eq_pzm1}
p(\mathbb{Z}|M) = \prod_{Z\in\mathbb{Z}} p(Z|M),
\end{equation}
where \(Z\) is a depth image or a single point cloud scan of LiDAR. Note that \(Z\) consists of many independent observations (\ie, depth image pixels or 3D points). We denote each of them as  $z \in Z$, then
\begin{equation}
\label{eq_pzm2}
p(Z|M) = \prod_{z\in Z} p(z|M).
\end{equation}
Assume that \(z\) follows a Gaussian distribution: 
 \begin{equation}
\label{eq_pzm_guass}
  p(z|M) \sim \mathcal{N}(z^{\prime}, \sigma^2)
 \end{equation}
where \(z^{\prime}\) and \(\sigma\) are the mean and covariance of the Gaussian distribution. \(z^{\prime}\) is produced by applying sensor model \(\mathcal{S}(\cdot)\) on \(M\) with the same observing configuration as \(z\):
\begin{equation}
\label{eq_sensor_model}
z^{\prime} = \mathcal{S}_z(M) =\mathcal{S}_z(\mathcal{M}, T_z),
\end{equation}
where  \(\mathcal{S}(\cdot)\) is a camera projection model for depth sensors and a range-bearing model for LiDARs. $T_z$ is the estimated sensor observing perspective for $z$. One can easily get $T_z$ from the estimated trajectory $\mathbb{T}$. For numerical stability, we take log on both sides of \eqref{eq_lr2}. We get a log-likelihood ratio (LLR)
\begin{equation}
\label{eq_llr1}
LLR = \log{p(\mathbb{Z}|M_1)} - \log{p(\mathbb{Z}|M_2)}.
\end{equation}
We define 
\begin{equation}
\label{eq_r}
r(M, \mathbb{Z}) =  -\log{p(\mathbb{Z}|M)}.
\end{equation}
When \(LLR>0\), which equivalents to \(r(M_1, \mathbb{Z}) < r(M_2, \mathbb{Z})\), \(M_1\) is more likely to be better than \(M_2\) in terms of dense map posterior probability.
This means that we can compare different \(r(M_i, \mathbb{Z})\) directly. The lower, the better. To further simplify \eqref{eq_r}, one can plug \eqref{eq_pzm1}, \eqref{eq_pzm2}, and \eqref{eq_pzm_guass} into \eqref{eq_r} to get
\begin{equation}
\begin{split}
\label{eq_pzm_expend}
-\log{p(\mathbb{Z}|M)} &= -\log \left[\prod_{Z\in\mathbb{Z}} \prod_{z\in Z} \frac{1}{{\sigma \sqrt {2\pi } }}e^{{{ - \left( {z - z^{\prime}} \right)^2 }  / 2\sigma ^2 }}\right]\\
&= \sum_{Z\in\mathbb{Z}} \sum_{z\in Z} \frac{\left( {z - z^{\prime}} \right)^2}{2\sigma ^2} + L_0,
\end{split}
\end{equation}
where \(L_0\) is a constant. Since \(L_0\) is the same for the same set of \(\mathbb{Z}\), then \eqref{eq_r} can be simplified to be
\begin{equation}
\label{eq_r_faster}
r(M, \mathbb{Z}) \sim \overline{r}(M, \mathbb{Z}) = \sum_{Z\in\mathbb{Z}} \sum_{z\in Z} \left( {z - z^{\prime}} \right)^2,
\end{equation}
where \(\overline{r}(M, \mathbb{Z})\) is the proposed metric DMP, the lower, the better. To calculate its value, one needs to provide a 3D reconstruction estimate $M$, sensor readings $\mathbb{Z}$, and a sensor model \(\mathcal{S}(\cdot)\).

\begin{figure*}[!hbt]\centering
\subfloat[fr1/desk]{ \includegraphics[width=0.194\linewidth]{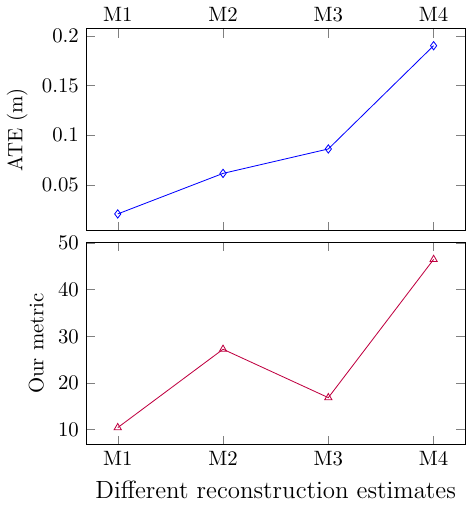}}
\subfloat[fr1/desk2]{ \includegraphics[width=0.19\linewidth]{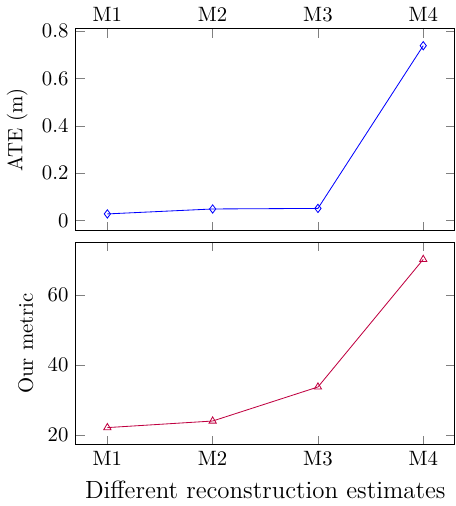}}
\subfloat[fr1/room]{ \includegraphics[width=0.19\linewidth]{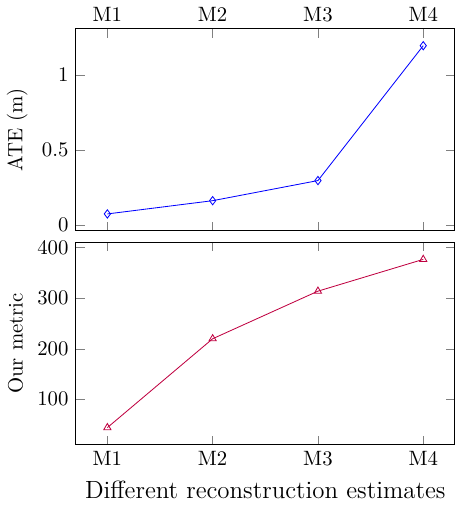}}
\subfloat[fr2/desk]{ \includegraphics[width=0.196\linewidth]{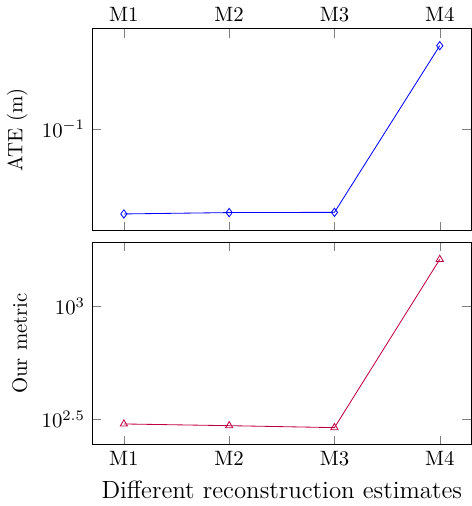}}
\subfloat[fr3/office]{ \includegraphics[width=0.196\linewidth]{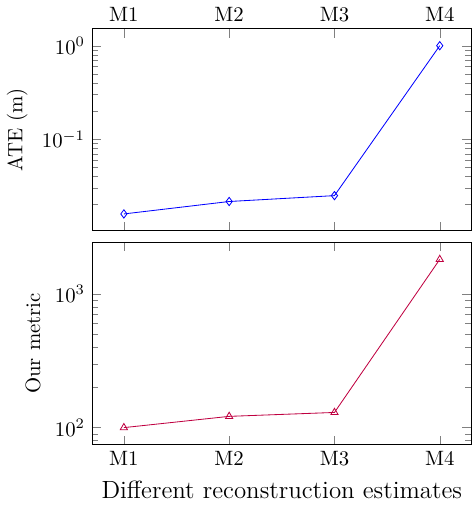}}
\caption{Comparison of ATE and our metric DMP. For each sequence, curves of two metrics follow almost the same ascending order, which means metrics have very similar comparison results. The outlier case in (a) is discussed in section \ref{sec_ineffective_cases}.}\label{fig:tum_eval_metric} % g checked
\end{figure*}

\section{Evaluation of DMP -- the Proposed Metric}
\label{sec:org521ea75}

To evaluate the proposed metric DMP, we conduct experiments to compare the results of different metrics on multiple datasets:  TUM RGB-D dataset~\cite{Sturm2012} and Augmented ICL-NUIM~\cite{Choi2015}. In all of our experiments, 3D models are fused using Surfels implemented by ElasticFusion~\cite{Whelan2015}.
\subsection{TUM RGB-D dataset}
\label{sec:org265c48e}

The TUM RGB-D dataset~\cite{Sturm2012}  is widely used for evaluating SLAM systems. The dataset has RGB-D sequences with ground truth camera trajectories available. We run experiments on a subset of sequences that are commonly used for SLAM evaluation~\cite{MurArtal2017}. Following \cite{Sturm2012}, we adapt the absolute trajectory error (ATE) as the baseline metric.  

For each data sequence, four 3D reconstruction estimates, $M_1$, $M_2$, $M_3$, and $M_4$ are used as benchmarking data. These estimates are from SLAM systems with different characteristics. The results are reported in Fig.~\ref{fig:tum_eval_metric}. For better visualization, $M_1$ to $M_4$ are sorted with the ATE metric. 

\noindent \textbf{Key observation.} In the results, the proposed metric DMP can report the same ascending order as the ATE metric in most cases. This means our metric can provide similar evaluation results as the ATE.

\subsection{Augmented ICL-NUIM dataset}
\label{sec:org859bf5e}
The augmented ICL-NUIM~(AIN) dataset~\cite{Choi2015} is a synthetic dataset with ground-truth surface models and camera trajectories. Experiments are performed to quantitatively compare performance of the proposed metric DMP with ground truth-based metrics: trajectory RMSE and surface mean distance (SMD).

\begin{figure}[!bht]\centering
\subfloat[livingroom 1]{ \includegraphics[width=0.39\linewidth]{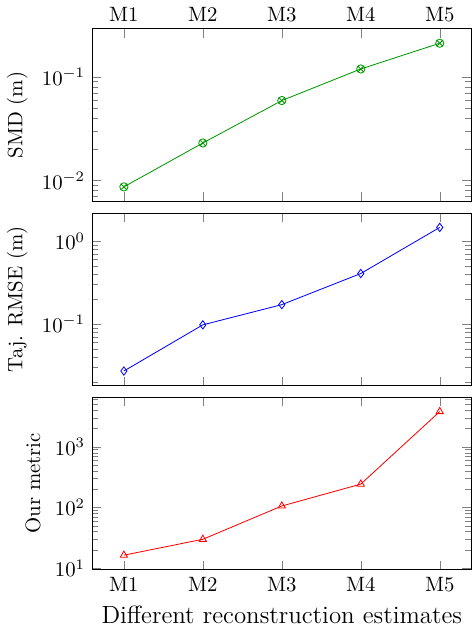}}
\subfloat[livingroom 2]{ \includegraphics[width=0.39\linewidth]{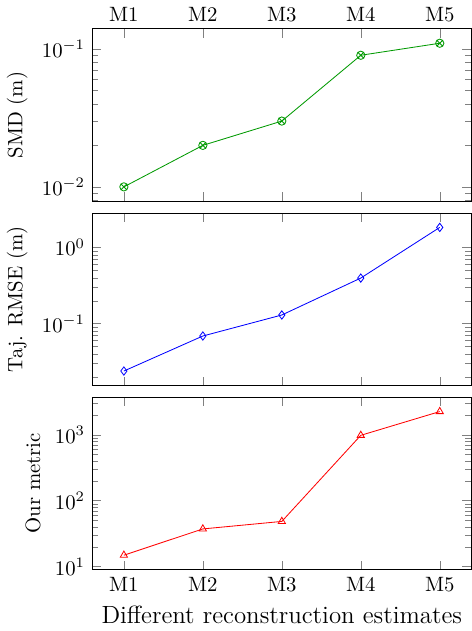}}\\
\subfloat[office 1]{ \includegraphics[width=0.40\linewidth]{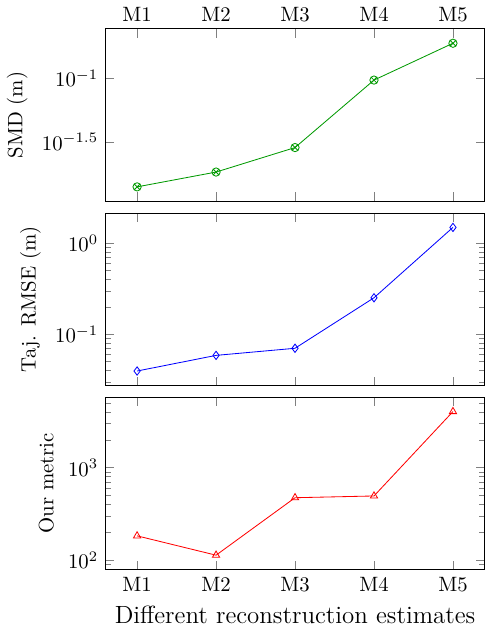}}
\subfloat[office 2]{ \includegraphics[width=0.39\linewidth]{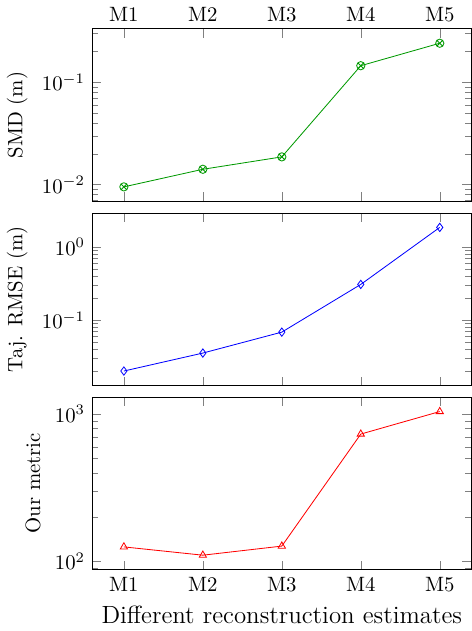}}\\
\caption{
Comparison of surface mean distance (SMD), trajectory RMSE (Traj. RMSE), and the proposed metric DMP. Our metric DMP positively correlates to SMD and Traj. RMSE, which means our metric can provide almost the same ranking results as ground truth-based metrics. The two outliers can be avoided by overlapping scans.
}\label{fig:nuim_eval_metric} % g checked
\end{figure}

\begin{figure}[!htb]\centering
\subfloat[$M_1$, DMP: 14.9]{ \includegraphics[width=0.35\linewidth]{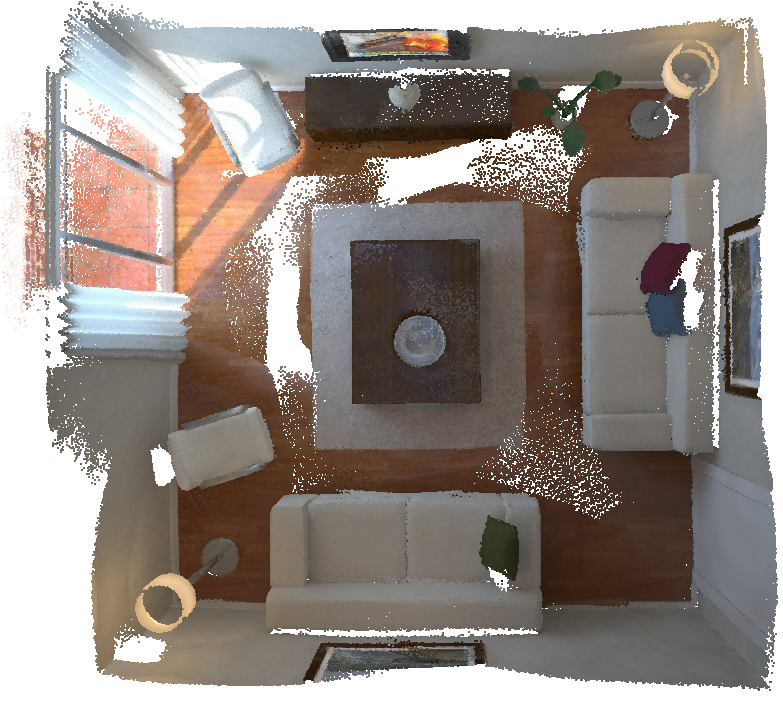}} % slam opted
\subfloat[$M_2$, DMP: 37.3]{ \includegraphics[width=0.35\linewidth]{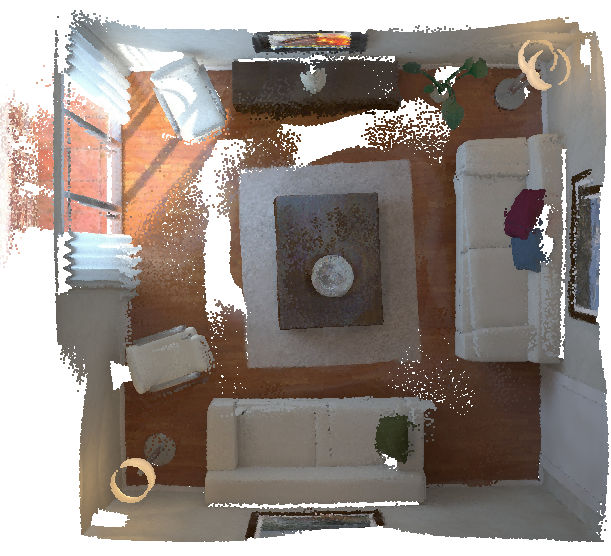}}\\ % odom
\subfloat[$M_3$, DMP: 48.5]{ \includegraphics[width=0.35\linewidth]{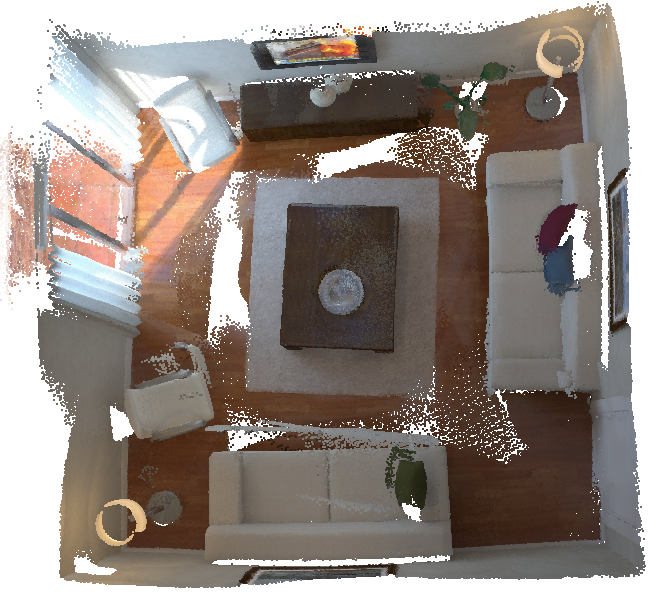}} % choi2015
\subfloat[$M_4$, DMP: 982.4]{ \includegraphics[width=0.35\linewidth]{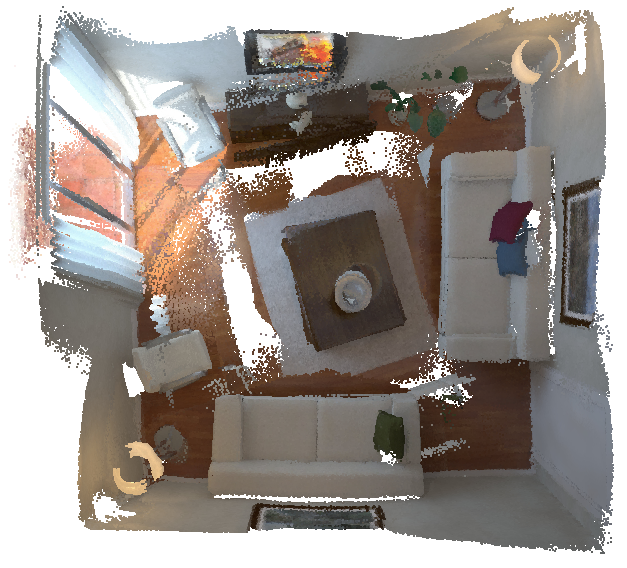}}\\
\subfloat[$M_5$, DMP: 2254.0]{ \includegraphics[width=0.5\linewidth]{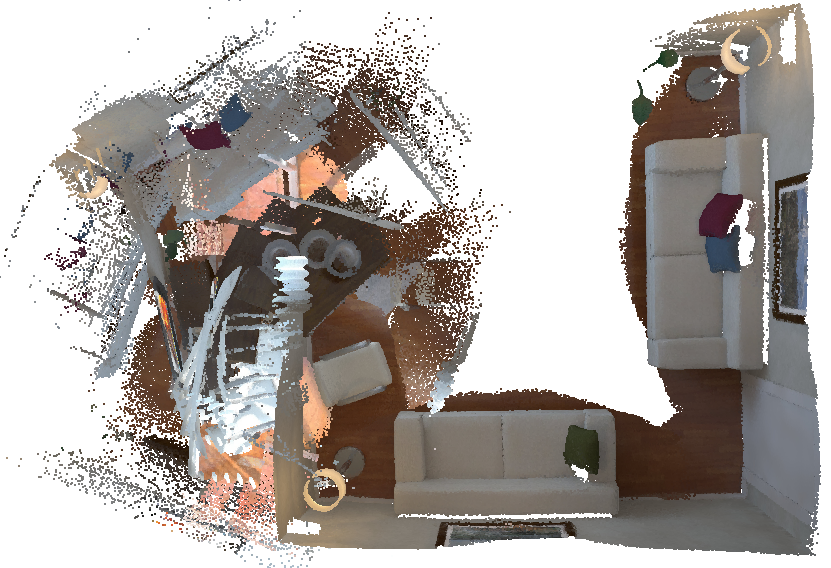}}
\caption{Visualization of estimated 3D models of the livingroom~2 sequence and their corresponding DMP values. In (a), $M_1$ leads to the best 3D mapping quality among the five estimates. There are not any noticeable mismatches. In (b), there are mismatches at the upper right-hand corner. The lamp and sofa are not well reconstructed. In (c), the sofa at the buttom of the image is not perfectly matched between frames. In (d), the coffee table at the center is off. (e) shows a collapsed 3D model optimized with a false positive loop.}\label{fig:nuim_metric_vis}
\end{figure}

Results are reported in Fig.~\ref{fig:nuim_eval_metric}. Reconstruction estimates data $M_1$-$M_5$ for evaluation are from SLAM methods and ranked with the SMD metric. \textbf{Key observations:} Our metric DMP positively correlates to SMD and Traj. RMSE, which means our metric can provide almost the same ranking results as ground truth-based metrics. To further visualize our metric, we include screen-shots of different 3D models and corresponding DMP values in Fig.~\ref{fig:nuim_metric_vis}.

\subsection{Ineffective cases}
% \label{sec:orge33361d}
\label{sec_ineffective_cases}
Among all the evaluations on the two datasets, there are rare cases where the proposed metric DMP gives different results from ground truth-based metrics: $M_3$ in \texttt{TUM/fr1/desk}, $M_2$ in \texttt{AIN/office 1}, and $M_2$ in \texttt{AIN/office 2}. We believe that it is because these data sequences have most of the space scanned only once without loopy coverage. That is the case where our metric cannot handle perfectly. However, it is simple to avoid this problem by adding overlapping scans.

\section{An application of DMP: Loop evaluation}
\label{sec:org134402d}
One important use case of the proposed metric DMP is to evaluate and filter loop detections for SLAM and 3D reconstruction systems. 
Because most loop detection systems do not have a perfect precision performance, which means false positive loop detections are inevitable. Meanwhile, SLAM and 3D reconstruction systems tend to be very sensitive to false-positive loops, which are likely to dramatically decrease the 3D model accuracy, as happened in Fig.~\ref{fig:nuim_metric_vis} (e).
With the proposed metric, it becomes possible to rank different loops and filter out the false positive ones. 

Again, we run experiments on the Aug ICL-NUIM dataset~\cite{Choi2015}, which also contains a point cloud registration benchmark operating on point cloud scene fragments.   \cite{Choi2015} also provides a set of point cloud registration results with high recall but low precision. These registration results can be used as a source of loop detections. A subset of the loops that are not connected by ORB-SLAM2~\cite{MurArtal2017} co-visibility graph are selected for our experiments. 

\begin{figure}[thb]\centering
\subfloat[livingroom 1]{ \includegraphics[width=0.34\linewidth]{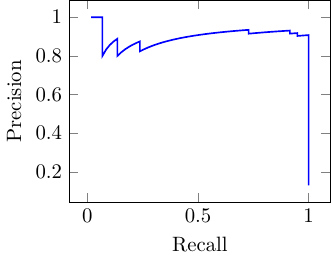}}
\subfloat[livingroom 2]{ \includegraphics[width=0.34\linewidth]{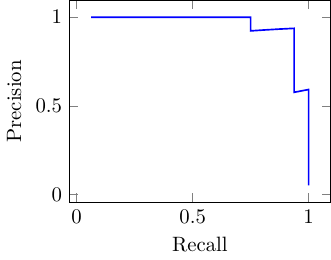}}\\
\subfloat[office 1]{ \includegraphics[width=0.34\linewidth]{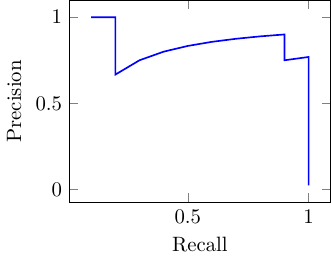}}
\subfloat[office 2]{ \includegraphics[width=0.34\linewidth]{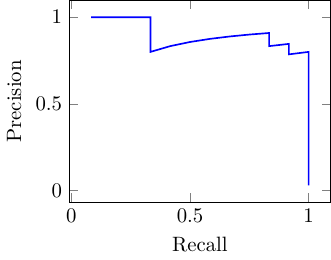}}\\
\caption{Precision and recall curves for ranking results based on the proposed metric DMP. The curves show a very good ranking performance.}\label{fig:nuim_PR}
\end{figure}

In our evaluation, we calculate a DMP value for each of the loops  by running the following steps:   
\begin{enumerate}
\item Perform pose graph optimization from ORB-SALM2 with a single loop added.
\item Fuse a 3D surfels model out of the optimization result.
\item Evaluate DMP.
\end{enumerate}
Finally, the loops are sorted by their DMP values. Precision-Recall curves, shown in Fig.~\ref{fig:nuim_PR}, are generated by varying threshold position. 

\noindent \textbf{Summary of observations.} In the results, we can see the curves all starts from \(100\%\) precision. Then the curves keep on high precision values when recall increases. There are a few drop points, which means false positive loops. But the number is limited. The majority of the false-positive loops are at the end of the list reflected by the sharp drops when recall reaches \(100\%\). These mean that the ranking has excellent performance, which proves that the proposed DMP is effective.

\section{Computational performance}
\label{sec:org3e827b2}
Our metric can be efficiently parallelized using OpenGL and CUDA. In our implementation, on an NVIDIA Titan X Pascal, the average evaluation time is 2.7 \(s\) for a TUM model and 4.2 \(s\) for an Augmented ICL-NUIM model. The speed can be further improved if only a sampled subset of data frames are used for evaluation.

\section{Conclusion}
\label{sec:orgca5ea1c}
In this work, we propose a metric, which can work without any ground truth data, for evaluating 3D reconstruction and mapping performance. In our experiments, the proposed metric DMP is benchmarked against ground truth based metrics. Results show that DMP can provide a similar evaluation capability.

The proposed metric not only makes 3D mapping evaluation simpler, but also opens many new opportunities. Our experiments show that it can evaluate loop detection results and lead to good precision-recall performance. We envision that more can be done with this metric, such as self-supervised methods and more available datasets.

\bibliographystyle{IEEEbib}
\bibliography{main.bib}
\end{document}